\documentclass[sigconf]{acmart}
\AtBeginDocument{%
  }

\copyrightyear{2025}
\acmYear{2025}
\setcopyright{acmlicensed}\acmConference[MMAsia '25]{ACM Multimedia Asia}{December 9--12, 2025}{Kuala Lumpur, Malaysia}
\acmBooktitle{ACM Multimedia Asia (MMAsia '25), December 9--12, 2025, Kuala Lumpur, Malaysia}
\acmDOI{10.1145/3743093.3771020}
\acmISBN{979-8-4007-2005-5/2025/12}

\acmSubmissionID{285}

\usepackage{multirow} 
\usepackage{makecell} 
\usepackage{siunitx}  
\usepackage{xcolor}   
\usepackage{hyperref}
\usepackage{graphicx}
\usepackage[ruled,vlined]{algorithm2e}

\renewcommand\footnotetextcopyrightpermission[1]{}

\begin{document}


\title{AVIR: Adaptive Visual In-Document Retrieval for Efficient Multi-Page Document Question Answering}

\author{Zongmin Li}
\affiliation{%
  \institution{China University of Petroleum (East China)}
  \city{Qingdao}
  \state{Shandong}
  \country{China}
}
\affiliation{%
  \institution{Shandong Xiehe University}
  \city{Jinan}
  \state{Shandong}
  \country{China}
}
\email{lizongmin@upc.edu.cn}

\author{Yachuan Li}
\authornote{Corresponding author}
\affiliation{%
  \institution{China University of Petroleum (East China)}
  \city{Qingdao}
  \state{Shandong}
  \country{China}
}
\affiliation{%
  \institution{Universitat Autonoma de Barcelona}
  \city{Cerdanyola}
  \state{Barcelona}
  \country{Spain}
}
\email{liyachuan@s.upc.edu.cn}

\author{Lei Kang}
\affiliation{%
  \institution{Universitat Autonoma de Barcelona}
  \city{Cerdanyola}
  \state{Barcelona}
  \country{Spain}
}
\email{lkang@cvc.uab.es}

\author{Dimosthenis Karatzas}
\affiliation{%
  \institution{Universitat Autonoma de Barcelona}
  \city{Cerdanyola}
  \state{Barcelona}
  \country{Spain}
}
\email{dimos@cvc.uab.cat}

\author{Wenkang Ma}
\affiliation{%
  \institution{China University of Petroleum (East China)}
  \city{Qingdao}
  \state{Shandong}
  \country{China}
}
\email{s23070027@s.upc.edu.cn}








\renewcommand{\shortauthors}{Zongmin Li et al.}

\begin{abstract}
 
Multi‑page Document Visual Question Answering (MP‑DocVQA) remains challenging because long documents not only strain computational resources but also reduce the effectiveness of the attention mechanism in large vision–language models (LVLMs).
We tackle these issues with an Adaptive Visual In‑document Retrieval (AVIR) framework. A lightweight retrieval model first scores each page for question relevance. 
Pages are then clustered according to the score distribution to adaptively select relevant content. The clustered pages are screened again by Top-K to keep the context compact.
However, for short documents, clustering reliability decreases, so we use a relevance probability threshold to select pages.
The selected pages alone are fed to a frozen LVLM for answer generation, eliminating the need for model fine‑tuning. 
The proposed AVIR framework reduces the average page count required for question answering by 70\%, while achieving an ANLS of 84.58\% on the MP-DocVQA dataset—surpassing previous methods with significantly lower computational cost.
The effectiveness of the proposed AVIR is also verified on the SlideVQA and DUDE benchmarks.
The code is available at \href{https://github.com/Li-yachuan/AVIR-main}{https://github.com/Li-yachuan/AVIR-main}.
\end{abstract}





\maketitle

\section{Introduction}
\label{sec:intro}

Multi‑page documents, such as financial reports, slide decks, contracts, and scientific papers, are pervasive in real‑world workflows. Enabling models to read these lengthy documents and answer open‑ended questions. Multi‑Page Document Visual Question Answering (MP‑DocVQA) is therefore a critical capability for modern Document‑AI systems.  

However, our experimental analysis reveals that while current Large Vision–Language Models (LVLMs) perform well on short, single-page inputs, they encounter two fundamental limitations when extended to long documents.
(1) Computation cost: Encoding dozens of pages with quadratic self‑attention quickly becomes prohibitive, making naïve end‑to‑end inference impractical for real‑time or resource‑constrained settings. (2) Context dilution: As a document grows, only a handful of pages contain information relevant to a given question, while the remaining pages introduce distracting or contradictory context that harms answer quality.

A natural remedy is to retrieve the most relevant pages first and run VQA only on that subset. Yet existing retrieval‑based pipelines still fall short because they often depend on heavy cross‑encoders whose overhead rivals the LVLM itself, and they rely on a rigid Top‑$K$ cutoff, so any retrieval error irrevocably removes the supporting page and cascades into a wrong answer.

This paper proposes an Adaptive Visual In-Document Retrieval (AVIR) framework that tackles both issues simultaneously. We design a lightweight Pix2Struct‑based retriever to score each page, followed by an adaptive page selector that analyzes the score distribution: for short documents, it applies threshold filtering; for longer ones, it clusters pages into relevant versus irrelevant groups and keeps the Top-K pages of the relevant pages. This dynamic strategy preserves recall while aggressively pruning noise, making the downstream LVLM, a quantized Qwen2.5‑VL, both faster and more accurate. Crucially, Qwen2.5‑VL remains frozen; prompt engineering alone provides strong cross‑domain generalization without costly fine‑tuning.

Our main contributions are threefold:
\vspace{-0.09cm}
\begin{enumerate}
    \item A systematic analysis showing how irrelevant pages impair current LVLMs and how lightweight retrieval restores efficiency.
    \item An adaptive page selector that mitigates retrieval errors by data‑driven thresholding and clustering, outperforming rigid Top‑$K$ baselines.
    \item A complete retrieval‑guided MP‑VQA pipeline that is efficient (only selected pages are processed) yet achieves superior performance on comprehensive public benchmarks of MP‑DocVQA, SlideVQA, and DUDE.
\end{enumerate}

\section{Related Work}
\label{sec:related}

Multi-page Document Visual Question Answering (MP-DocVQA) enables models to comprehend and respond to questions based on complex, multi-page documents. Due to challenges such as lengthy content, intricate layouts, and cross-page dependencies, conventional single-page VQA methods exhibit limited effectiveness in this domain. Existing approaches to MP-VQA typically fall into three categories: sequence-based methods, structured fusion approaches, and retrieval-based frameworks.

\subsection{Sequence-based MP-VQA}

Sequence-based methods treat multi‑page documents as \textit{flattened sequences} and apply Transformer architectures originally devised for long‑form language modelling. Early work such as Longformer~\cite{beltagy2020longformer} and BIGBIRD~\cite{zaheer2020big} extends local or sparse global attention to accommodate thousands of tokens. RM‑T5~\cite{dong2024multi} introduces a recurrent memory cache so that a fixed‑size core iteratively processes page chunks and accumulates context. GRAM~\cite{blau2024gram} pushes the idea further by adding global reasoning tokens that attend to every page token, enabling holistic cross‑page aggregation within a single encoder–decoder pass. Arctic‑TILT~\cite{borchmann2024arctic} revisits the TILT architecture with byte‑level tokenisation and lightweight sparse attention, scaling end‑to‑end processing up to 400k tokens on commodity GPUs. Such models eliminate explicit page segmentation but still incur heavy memory footprints on very long documents.

\subsection{Structured Fusion MP-VQA}


Structured fusion methods explicitly model the hierarchical, spatial, and semantic relationships inherent in multi-page documents to better capture their complex structure. For instance, Hi-VT5\cite{tito2023hierarchical} extends sequence-to-sequence models with a hierarchical encoder-decoder that processes page tokens at multiple granularities, enabling effective cross-page information integration. DocFormer-v2\cite{appalaraju2024docformerv2} employs multi-window two-dimensional attention combined with squeezed layout tokens to achieve fine-grained layout reasoning and spatial understanding. These approaches build structured representations that jointly capture visual and textual relationships critical for multi-page document understanding. Despite their strong performance, they often introduce significant computational overhead, which motivates the development of lighter and more efficient methods.

\subsection{Retrieval-based MP-VQA}


Retrieval-based approaches decompose multi-page visual question answering (MP-VQA) into two distinct stages: first, retrieving the most relevant document pages, and second, performing question answering on the selected pages. Typically, existing methods rely on a straightforward Top-K page selection strategy, where only the top-ranked pages based on retrieval scores are passed to the reader model. 
M3DOCRAG\cite{cho2024m3docrag} improves retrieval quality by employing more sophisticated retrieval models that better capture relevance across pages, but these improvements often come with increased computational costs. More recently, SelfAttnScoring\cite{kang2024multi} introduces lighter-weight retrieval modules designed to enhance efficiency. Despite these gains, it still depends on a fixed Top-K selection mechanism, which leaves it vulnerable to the same retrieval error issues and limits its robustness.

\vspace{0.5cm}
Thus, our proposed method addresses the limitations of existing retrieval-based frameworks through an \textbf{adaptive in-document retrieval guidance approach}, effectively balancing efficiency and accuracy. Specifically, we introduce a lightweight retrieval module coupled with a novel adaptive page selector, dynamically classifying relevant pages based on the distribution of retrieval scores. This adaptive mechanism significantly mitigates the impact of retrieval errors, reduces irrelevant context processing, and greatly enhances computational efficiency.

\section{Proposed Methodology}
\label{sec:method}

\subsection{Question Definition}
\label{sec:method-qd}
\begin{figure}
    \centering
    \includegraphics[width=\linewidth]{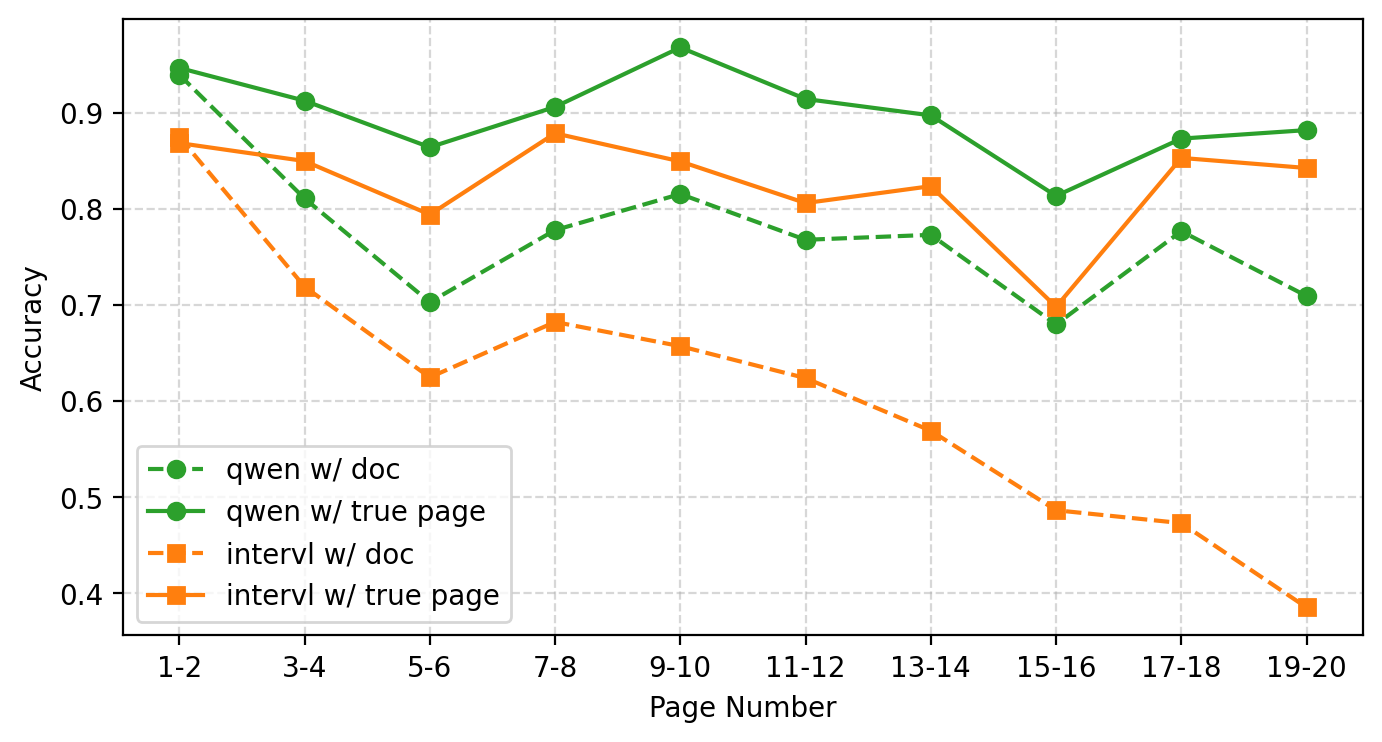}
    \caption{The influence of document length on the accuracy of MP-VQA. The experiment is conducted on the validation set of MP-DocVQA, where the answers are usually on a single page. 'w/ doc' indicates using the entire document, while 'w/ true page' means only the correct page is used. Due to the uneven distribution of page lengths, we have only selected the first 202 documents for each category.}
    \label{fig:lenth_acc}
\end{figure}

Although current large-scale visual-language models (LVLMs) can directly handle multi-page documents, their performance degrades significantly as the document length increases. We evaluate this problem on the MP-DocVQA validation set using Qwen2.5-VL (3B) and interVL3 (2B) as examples, and the results are shown in Fig.~\ref{fig:lenth_acc}.

When the document contains only 1-2 pages, both models achieve high accuracy when using the entire document or only relevant pages. However, when the document length exceeds 3 pages, the performance of both models degrades significantly when processing the entire document. The significant performance gap between the full document setting and the ground truth page setting confirms that irrelevant context introduces significant noise.

The performance degradation of interVL3 worsens with increasing document length, especially when the document length exceeds 8 pages. Although Qwen2.5-VL shows higher robustness than interVL3, its performance on longer documents is still significantly degraded compared to only relevant pages Q\&A.

Despite the differences between the two models, they both show a consistent trend: LVLM performs significantly worse on long documents than on short documents. This highlights the challenges that current models face in handling long-range dependencies and context clutter in multi-page settings.

\begin{figure*}
    \centering
    \includegraphics[width=\linewidth]{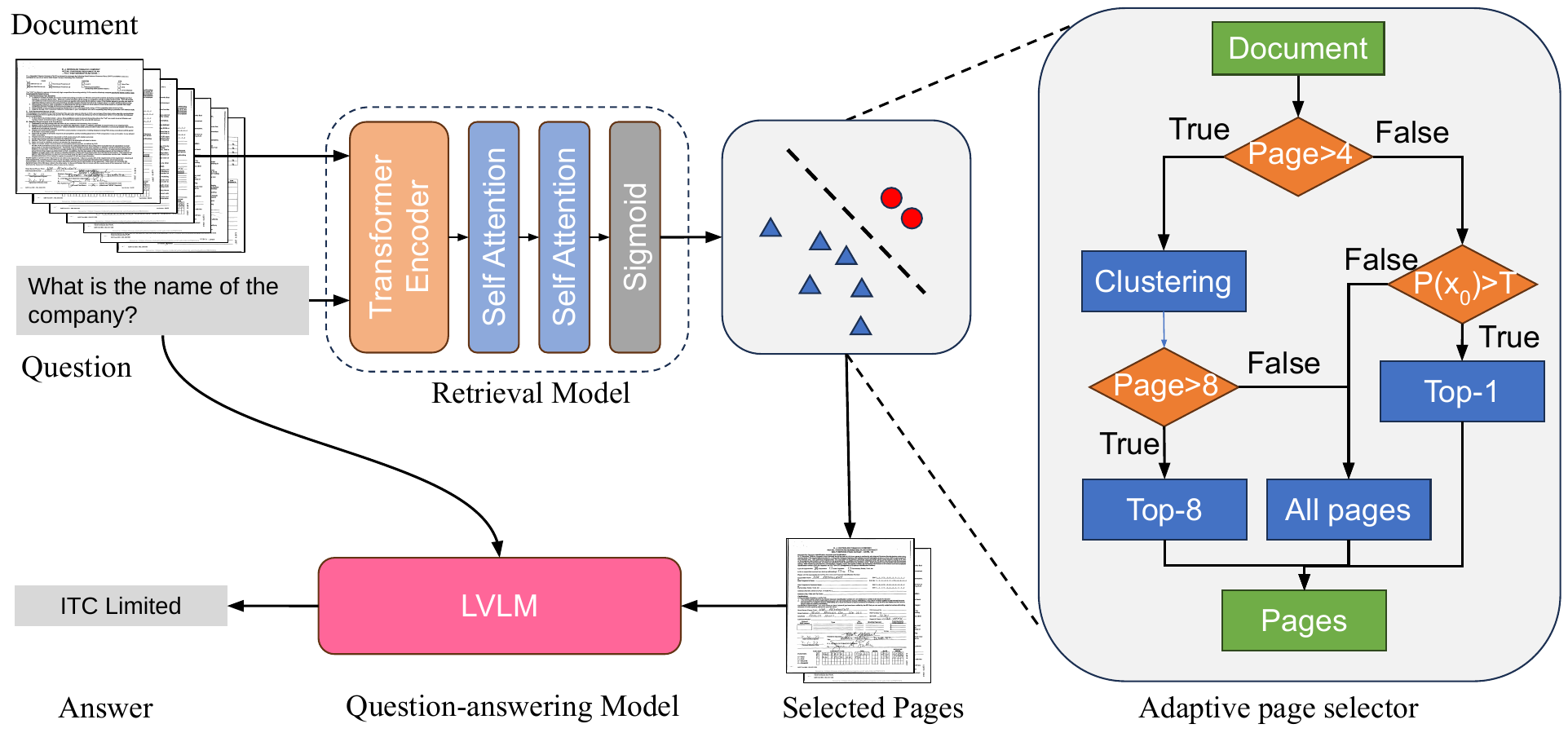}
    \caption{Overview of the proposed AVIR. $P(x_0)$ means The relevance probability assigned by the retrieval model to the most relevant page.}
    \label{fig:stru}
\end{figure*}

\subsection{Overview}

To alleviate the performance degradation of large models on long documents, retrieval-augmented generation (RAG) has emerged as a promising paradigm. However, existing RAG-based methods are highly sensitive to retrieval errors, which can lead to significant performance degradation.

To mitigate the impact of retrieval errors and enhance overall performance, we propose a novel adaptive visual in-document retrieval framework, termed AVIR, tailored for multi-page document question answering (MP-DQA), as illustrated in Fig.~\ref{fig:stru}. Our method follows the general RAG pipeline but introduces key innovations to improve retrieval robustness and answer accuracy. Specifically, the framework consists of three sequential components:

\textbf{Page Retrieval}: We employ a page-level retriever that estimates the semantic relevance between each document page and the given question.
\textbf{Adaptive Page Selection}: Based on the initial relevance scores, an adaptive page selector—designed to consider both score distribution and inter-page relationships—is introduced to select the most informative page(s) for downstream reasoning.
\textbf{Answer Generation}: Finally, the selected pages and the question are fed into a large pre-trained LVLM for answer generation. To balance accuracy and computational efficiency, we adopt Qwen2.5-VL~\cite{bai2025qwen2_5} as our backbone LVLM.

In the following sections, we provide a detailed introduction to the page retrieval module and our adaptive page selector. The architecture can significantly reduce retrieval noise and enhance the performance of long-document VQA systems.

\subsection{Page Retrieval Model}

An essential component of retrieval-augmented generation (RAG) frameworks is the page retrieval module, which identifies the most relevant pages from a multi-page document in response to a given question. While several high-performance retrieval models have been proposed—such as ColBERTa-based architectures like ColPali~\cite{faysse2024colpali} and document structure-enhanced models like DSE~\cite{ma2024unifying}—they typically require substantial computational resources. This becomes particularly problematic in the multi-page setting, where the retrieval model must evaluate every individual page in the document. In some cases, the computational cost of such large-scale retrieval models even exceeds that of the answer generation module itself.

To ensure the end-to-end efficiency and practical applicability of the framework, we adopt a lightweight and efficient page retrieval model. Inspired by SelfAttnScoring~\cite{kang2024multi}, we leverage the encoder of Pix2Struct~\cite{lee2023pix2struct}, a visual language model designed for document understanding, to extract visual semantic features from each page. On top of the frozen encoder, a custom relevance scoring head is built, consisting of two self-attention layers and a sigmoid activation function that maps page-question feature alignments to relevance probability space.

Our retrieval module contains only approximately 100 million parameters, which is nearly 4\% the size of the downstream LVLM used for question answering. While this lightweight design inevitably limits the model’s generalization capability, requiring task-specific fine-tuning on each dataset, we find this trade-off acceptable. In practice, the gain in inference efficiency significantly outweighs the one-time fine-tuning cost, especially when deploying the system at scale.

\subsection{Adaptive Page Selection}

The adaptive page selector serves as a critical component of our proposed framework. It is specifically designed to address three major challenges in multi-page document question answering (MP-DQA):

\begin{enumerate}
    \item Mitigating the performance degradation of large-scale LVLMs on long documents, as discussed in Section~\ref{sec:method-qd};
    \item Improving computational efficiency by reducing the number of pages passed to the answer generation module;
    \item Reducing the negative impact of retrieval errors introduced during the initial page relevance estimation stage.
\end{enumerate}

To this end, we propose a dynamic selection strategy that adapts to the number of pages in a document and the distribution of their relevance scores.

For longer documents (i.e., those exceeding four pages), we first employ a relevance-based clustering strategy to categorize pages into two groups: relevant and irrelevant. If the number of relevant pages exceeds a predefined threshold (set to 8), we apply a hard cutoff by selecting the top 8 pages with the highest relevance scores.
This step is essential for preserving answer quality, as supplying too many pages to the LVLM not only introduces noise but also increases the reasoning complexity, which can negatively impact performance.
Empirical results show that the top 8 most relevant pages are typically sufficient to encompass the information necessary to answer the vast majority of questions.

While the clustering strategy works well for long documents, it is less suitable for short documents (fewer than four pages), where forced binary classification may lead to the exclusion of relevant pages with marginally lower relevance scores.
For instance, if a three-page document yields relevance probabilities of 0.34, 0.33, and 0.33, clustering would retain only the first page, despite all three being similarly relevant. Such close score distributions are more common in short documents.
As shown in Fig.~\ref{fig:lenth_acc}, the question-answering performance on short documents is generally less sensitive to the number of included pages. Therefore, we adopt a threshold-based filtering strategy for this case.
Specifically, if any page has a relevance score above a predefined threshold (set to 0.6 in our experiments), only those pages are retained; otherwise, all pages are preserved.
This simple yet effective approach helps prevent mistakenly discarding relevant pages in short documents, thereby improving the robustness and error tolerance of the overall system.

In summary, this adaptive selection mechanism strikes a balance between retrieval robustness and computational efficiency. It ensures that highly relevant information is retained while excluding irrelevant or marginally relevant content, thereby enhancing the reliability and effectiveness of the downstream answer generation process.
The pseudo code of the adaptive page selector is shown in Algorithm~\ref{alg:page-selector}.

\begin{algorithm}[htbp]
\caption{Adaptive Page Selector with K-Means Clustering}
\label{alg:page-selector}
\KwIn{Document pages $P = \{p_1, p_2, \dots, p_n\}$; \\
\quad\quad\quad Relevance scores $R = \{r_1, r_2, \dots, r_n\}$; \\
\quad\quad\quad Threshold $T = 0.6$; Max selected pages $K = 8$}
\KwOut{Selected page subset $P_{\text{selected}}$}

\If{$n < 4$}{
    \If{exists $i$ such that $r_i \geq T$}{
        $P_{\text{selected}} \gets \{p_i \mid r_i \geq T\}$ \;
    }
    \Else{
        $P_{\text{selected}} \gets P$ \tcp*{Select all pages}
    }
}
\Else{
    Apply K-Means clustering ($k=2$) on $R$ to divide pages into: \\
    \quad $C_{\text{rel}}$ (relevant cluster), $C_{\text{irrel}}$ (irrelevant cluster) \;

    Let $P_{\text{rel}}$ be pages in $C_{\text{rel}}$ \;

    \If{$|P_{\text{rel}}| > K$}{
        Sort $P_{\text{rel}}$ by relevance score in descending order \;
        $P_{\text{rel}} \gets$ top-$K$ pages in $P_{\text{rel}}$ \;
    }

    $P_{\text{selected}} \gets P_{\text{rel}}$ \;
}
\Return{$P_{\text{selected}}$}
\end{algorithm}

\section{Experiments}
\label{sec:experiments}

\subsection{Dataset and Metrics}

We evaluate our model on three multi-page document visual question answering (MP-VQA) datasets, each designed to capture different challenges associated with understanding long-form documents: SlideVQA, MP-DocVQA, and DUDE.

\textbf{SlideVQA}~\cite{tanaka2023slidevqa} contains presentation-style documents averaging 20 pages. Questions reference content spread across multiple slides, requiring sequential, layout-aware integration of visual and textual information. Evaluation uses Exact Match (EM) accuracy and token-level F1 score, emphasizing temporal reasoning and structured document understanding.
\textbf{MP-DocVQA}\cite{tito2023hierarchical}, built on DocVQA\cite{mathew2021docvqa}, focuses on multi-page documents averaging 8.3 pages with dense content. Questions often span spatially distant regions. Evaluation metrics include Accuracy and ANLS~\cite{biten2019scene}, covering various formats such as manuals, reports, and forms.
\textbf{DUDE}~\cite{van2023document} comprises 3,000 real-world PDFs (up to 25 pages) across finance, healthcare, and law. It includes 23.7K questions of four types: Extractive, Abstractive, List, and Unanswerable. Evaluated with ANLS, it introduces domain diversity and reasoning complexity, and offers fine-grained performance analysis by question type.

\subsection{Implementation details}

We use PyTorch, Transformers, and FlashAttention2 libraries for running models. The minimum and maximum pixels of LM are $768 \times 28 \times 28$ and $5120 \times 28 \times 28$, which follow the setting of Qwen2.5~\cite{bai2025qwen2_5}. All experiments are conducted with a single A40 46GB GPU. 

We adopt the pre-trained Qwen2.5-VL-3B~\cite{bai2025qwen2_5} as the question-answering model, without any task-specific fine-tuning. To improve inference efficiency, we use its AWQ-quantized version, denoted as Qwen2.5-VL-3B-AWQ.
Since the adaptive page selector is non-parametric, the only trainable component is the page retrieval model, which contains approximately 0.1B parameters and is trained following the SelfAttnScoring strategy~\cite{kang2024multi}.

\subsection{Comparison with the State of the Art}

\subsubsection{Results on MP-DocVQA Dataset}

\begin{table}[h]
\centering
\caption{Comparison with state of the art on the test set of MP-DocVQA dataset.}
\label{tab:comparison}
    \setlength{\tabcolsep}{2pt}
\begin{tabular}{l c c c c c}
\hline
Method & Year & OCR & Params. & Page Pred. (\%) & ANLS \\
\hline
BERT \cite{devlin2019bert} & 2018 & \checkmark & 334M & 71.24 & 0.5347 \\
T5 \cite{raffel2020exploring} & 2020 & \checkmark & 223M & 46.05 & 0.4028 \\
Longformer \cite{beltagy2020longformer} & 2020 & \checkmark & 148M & 70.37 & 0.5506 \\
Big Bird \cite{zaheer2020big} & 2020 & \checkmark & 131M & 72.27 & 0.5854 \\
LayoutLMv3 \cite{huang2022layoutlmv3} & 2022 & \checkmark & 125M & 74.02 & 0.5513 \\
Hi-VT5 \cite{tito2023hierarchical} & 2023 & \checkmark & 316M & 79.23 & 0.6201 \\
ScreenAI~\cite{baechler2024screenai} &2024&\checkmark&5B&77.88&0.7711\\
GRAM~\cite{blau2024gram} &2024&\checkmark&859M&-&0.8032\\ 
Arctic-TILT~\cite{borchmann2024arctic}&2024&\checkmark&800M&-&0.8122\\
RMT5~\cite{dong2024multi}& 2024&\checkmark&312M&\textbf{88.32}&0.6401\\

SelfAttnScoring~\cite{kang2024multi} & 2024 & - & 273M & 81.55 & 0.6199 \\
M3DOCRAG~\cite{cho2024m3docrag}&2024&-&8B&81.05&0.8444\\
Qwen2.5-3B-AWQ~\cite{bai2025qwen2_5}&2025&-&3B&-&0.8405\\
\textbf{AVIR (Ours)} & - & - & 3B & 81.55 & \textbf{0.8458} \\
\hline
\end{tabular}
\end{table}

We systematically compare our method with both OCR-based and OCR-free approaches, as summarized in Table~\ref{tab:comparison}. It is evident that earlier methods relied heavily on OCR to extract textual content, while recent advances in large vision-language models (LVLMs) have enabled OCR-free methods to achieve comparable or even superior performance.
Our method achieves an ANLS score of 0.8458, outperforming many strong OCR-based baselines and surpassing existing OCR-free methods in accuracy. Notably, our model maintains a relatively small parameter size. Although ARIV is larger than recent lightweight models like GRAM and Arctic-TILT, it is significantly smaller than large-scale models such as ScreenAI and M3DOCRAG.
A key advantage of our approach lies in its retrieval-augmented design. The retrieval module contains only about 100 million parameters, and the question-answering model processes only a small subset of the most relevant retrieved pages. This drastically reduces the overall computational overhead, making our method both accurate and efficient.
Our baseline QA model, Qwen2.5-3B-AWQ, is a fully end-to-end LVLM with fewer parameters than our method. However, due to the need to encode the entire document, its actual computation cost is significantly higher.
Among all baselines, M3DOCRAG is closest in accuracy, trailing ours by only 0.14\% ANLS. However, this marginal improvement comes at the expense of a larger model size, with a total number of parameters nearly 3 times that of the proposed AVIR.
Furthermore, since both its retrieval and QA components rely on large models, its computational efficiency is substantially lower.
In terms of page retrieval accuracy, our method achieves the same Top-1 accuracy as SelfAttnScoring. However, our retrieval mechanism is more tolerant to ranking noise, making it more robust: errors from the retriever have a smaller negative impact on final performance. We provide a more detailed analysis of efficiency and ablations in Section~\ref{sec:ablation}.

\subsubsection{Results on SlideVQA Dataset}

\begin{table}[htbp]
    \centering
    \caption{Performance on the SlideVQA benchmark. “T/L/V” denotes the “text/layout/visual” modality of images.}
    \label{tab:slidevqa_results}
    \setlength{\tabcolsep}{4pt}
    \begin{tabular}{lcc cc c}
        \hline
        Model&Year&Modal&Params&EM & F1 \\
        \hline
        T5~\cite{raffel2020exploring}&2020 & T & 0.2B & 29.3 & 37.9 \\
        LayoutT5~\cite{tanaka2021visualmrc}&2021 & TLV&-& 31.7 & 39.9 \\
        LayoutLMv2~\cite{xu2020layoutlmv2}& 2020&TLV &-& 21.4 & 29.3 \\
        FID~\cite{izacard2020leveraging} &2020& T&-&30.4 & 38.9 \\
        M3D~\cite{tanaka2023slidevqa} &2023& TLV&-&33.5 & 41.7 \\
        BLIP-2~\cite{li2023blip} &2023&TV&3.4B &28.3 & 38.8 \\
        InstructDr~\cite{tanaka2024instructdoc}&2024&TLV & 3.4B& 31.9 & 40.2 \\
        Arctic-TILT~\cite{borchmann2024arctic}&2024&V&0.8B&55.1 &-\\
        VDocRAG~\cite{tanaka2025vdocrag}&2025& V &8B &- & 44.2\\
        FRAG~\cite{huang2025frag}       &2025&V& 7B  &59.8&65.1\\
        Qwen2.5-3B-AWQ~\cite{bai2025qwen2_5}&2025&V&3B&56.6&65.8\\
        Eagle-2.5~\cite{chen2025eagle}&2025&V &8B&\textbf{63.2} & \textbf{72.3} \\
        \textbf{AVIR (Ours)}&-&V &3B& 60.3 & 68.9 \\
        \hline
    \end{tabular}
\end{table}

SlideVQA is a slide-based document dataset characterized by rich layout information and visually structured content. Questions in SlideVQA often require reasoning over content that is scattered across multiple slides, making cross-page understanding a critical capability. Early methods typically integrate textual, layout, and visual features to address this challenge.
The evaluation results on the SlideVQA test set are presented in Table~\ref{tab:slidevqa_results}. Although earlier approaches such as LayoutT5 and InstructDr incorporate multiple modalities (text, layout, and vision), their performance has been surpassed by more recent pure vision-based models. Our method also relies solely on visual input and achieves strong results, with an Exact Match (EM) score of 60.3 and an F1 score of 68.9, outperforming our baseline Qwen2.5-3B-AWQ by 3.7 and 3.1 points, respectively. These results are second only to Eagle-2.5. 
Eagle-2.5 achieves higher scores, but at the cost of a significant increase in model size—nearly 3 times that of AVIR.
Importantly, our retrieval-based approach, AVIR, avoids processing the entire document directly with a large-scale VQA model, further speeding up inference and reducing computational cost.
Overall, the results on SlideVQA further demonstrate the effectiveness and generalizability of our retrieval-augmented approach for long-document VQA, highlighting its ability to scale across document types with varying layout complexity and cross-page dependencies.

\subsubsection{Results on DUDE Dataset}

\begin{table*}[htbp]
    \centering
    \caption{Comparison of Models on the DUDE benchmark.}
    \label{tab:dude_results}
    \begin{tabular}{llcccc}
        \hline
        \multirow{2}{*}{\textbf{Method}} & \multirow{2}{*}{\textbf{ANLS}} & \multicolumn{4}{c}{\textbf{ANLS per Answer Type}} \\
        \cline{3-6}
        & & Extractive & Abstractive & List-of-answers & Unanswerable \\
        \hline
        Arctic-TILT~\cite{borchmann2024arctic} & \textbf{0.5809} & 0.6271 & 0.5645 & 0.4669 & \textbf{0.6261} \\
        GPT-4\&Azure OCR & 0.5392 & 0.5973 & 0.5248 & \textbf{0.5785} & 0.5131 \\
        GRAM~\cite{blau2024gram} & 0.5336 & 0.5683 & 0.5232 & 0.1996 & 0.6543 \\
        GRAM C-Former~\cite{blau2024gram} & 0.5097 & 0.5515 & 0.5046 & 0.1726 & 0.6104 \\
        DocGptVQA & 0.5002 & 0.5186 & 0.4832 & 0.2822 & 0.6204 \\
        \textbf{AVIR (Ours)} & 0.4905 & \textbf{0.6754} &\textbf{ 0.6404 }& 0.1029 & 0.0000 \\
        DocBlipVQA & 0.4762 & 0.5069 & 0.4631 & 0.3073 & 0.5522 \\
        \textbf{Qwen2.5-3B-AWQ (Our baseline)}~\cite{bai2025qwen2_5} & 0.4575 & 0.6319 & 0.5940 & 0.1199 & 0.0000 \\
        T5-concat & 0.3867 & 0.3727 & 0.3750 & 0.1681 & 0.5289 \\
        Multi-Modal T5 VQA & 0.3790 & 0.4155 & 0.4024 & 0.2021 & 0.3467 \\
        HI-VT5 \cite{tito2023hierarchical}  & 0.3574 & 0.2831 & 0.3298 & 0.1060 & 0.6290 \\
        QAP & 0.1159 & 0.0009 & 0.0007 & 0.0000 & 0.6199 \\
        \hline
    \end{tabular}
\end{table*}

\begin{figure}
    \centering
    \includegraphics[width=1\linewidth]{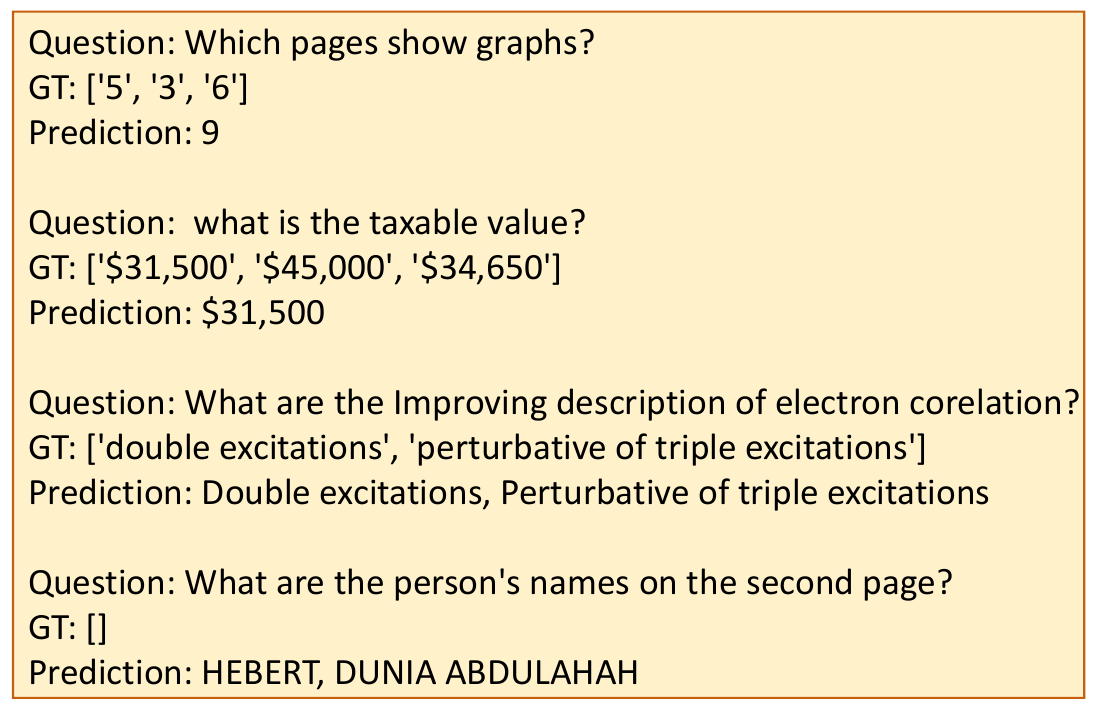}
    \caption{Four typical errors of AVIR on the DUDE dataset.}
    \label{fig:error_dude}
\end{figure}

The DUDE dataset presents a unique challenge that is not directly compatible with our method, as its questions involve four distinct answer types—Extractive, Abstractive, List-of-answers, and Unanswerable—which require models to handle diverse output formats. Without targeted fine-tuning, models struggle to generalize across such heterogeneous formats. As shown in Table~\ref{tab:dude_results}, our method, along with the baseline Qwen2.5-3B-AWQ, achieves near-zero performance on the List-of-answers and Unanswerable categories, while all other listed methods are fine-tuned specifically on DUDE. 
We show four typical mistakes of AVIR in Fig.~\ref{fig:error_dude}: 1) The question-answering model cannot handle global questions because it does not access all pages. 2) The question answer is not comprehensive and cannot find all answers. 3) Although the answer is correct, it cannot be answered in the specified format. 4) Nonsense in questions without answers.

Nevertheless, for the Extractive and Abstractive types—whose formats are commonly shared with other datasets—our method achieves state-of-the-art performance, outperforming the baseline by approximately 0.04 ANLS. This result highlights the effectiveness of our retrieval-augmented design even in DUDE. During evaluation, we employ a general prompt: "Answer the question using a single word or phrase." When this is modified to match DUDE’s expected format, e.g., "List all answers in one JSON array without anything else. Use 'none' for unanswerable questions," the performance on List-of-answers and Unanswerable questions does not improve and instead causes a significant drop in performance on the Extractive and Abstractive categories. These findings suggest that DUDE is not well-suited for the Qwen-2.5-vl model without fine-tuning, due to its strict and diverse answer format requirements. Nonetheless, the strong results on the Extractive and Abstractive answers demonstrate the robustness and generalizability of the proposed AVIR.

\subsection{Ablation experiment}
\label{sec:ablation}
We conduct ablation experiments to investigate the impact of retrieval on MP-DocVQA performance. The results are reported in Table~\ref{tab:rtr_eff}.
We can observe that in the baseline, where we ask questions directly without using retrieval, the question answering model not only inefficiently processes all pages but also suffers from a large number of irrelevant pages, resulting in low accuracy. This is illustrated in Fig.~\ref{fig:lenth_acc}. Following the SelfAttnScoring~\cite{kang2024multi}, using only the top-1 relevant pages improves the efficiency of the question answering model, but retrieval errors significantly degrade performance. As the number of Top-K pages increases, model performance gradually improves, reaching peak performance at Top-4, the strategy used by M3DOCRAG~\cite{cho2024m3docrag}. However, further increasing the Top-K results in a decrease in question answering performance. When Top-K reaches 20 (the maximum number of pages in the SlideVQA dataset), the model degenerates to non-retrieval question answering (our baseline).

Compared to the baseline, the proposed AVIR method effectively reduces interference from irrelevant pages. In contrast to the Top-K approach, it also alleviates the negative impact of retrieval errors, leading to optimal performance.
Moreover, our method requires only an average of 2.9 pages to answer each question, striking an ideal balance between effectiveness and efficiency.


\begin{table}
\caption{Ablation experiment on SlideVQA dataset. APS means the proposed Adaptive Page Selector. }
\label{tab:rtr_eff}
\setlength{\tabcolsep}{4pt}
\centering
\begin{tabular}{cccc}
\hline
\textbf{Method} & Ave. page & EM & F1 \\
\hline
Qwen2.5-VL-AWQ (baseline) & 20.0 & 56.6 & 65.8 \\
Retrival+Top-K (K=1) & 1.0 & 52.7 & 60.2 \\
Retrival+Top-K (K=2) & 2.0 & 56.5 & 64.6 \\
Retrival+Top-K (K=4) & 4.0 & 58.3 & 66.8 \\
Retrival+Top-K (K=8) & 8.0 & 57.2 & 65.7 \\
Retrival+APS (AVIR) & 2.9 & 60.3 & 68.9 \\
\hline
\end{tabular}
\end{table}

\section{Conclusions}

In this paper, we propose an Adaptive Visual In-Document Retrieval (AVIR) for efficient
multi-page document question answering, which effectively alleviates the problems of low computational efficiency and low accuracy of long documents in large visual-language models. The proposed adaptive page selector can effectively select highly relevant pages and alleviate the negative impact of retrieval errors. The experimental results on multiple datasets show that AVIR surpasses the baseline in both accuracy and efficiency and achieves the best performance among models of similar scale.

\textbf{Limitation:} Our method is not able to handle non-universal answer formats, such as those found in DUDE, unless the question answering model is specifically fine-tuned to do so. Moreover, because our method leverages a retrieval mechanism that only exposes a subset of relevant pages to the question answering model, it lacks access to the full document context. As a result, it is inherently incapable of addressing global questions that require complete document-level understanding, even for seemingly simple queries like, “How many pages does the document have?”.

\begin{acks}
This work is supported by the China Scholarship Council (CSC) (Grant no. 202406450075), the National key R\&d program (Grant no. 2019YFF0301800), the National Natural Science Foundation of China (Grant no. 61379106), and the Shandong Provincial Natural Science Foundation (Grant nos.ZR2013FM036, ZR2015FM011). 
\end{acks} 

\bibliographystyle{ACM-Reference-Format}
\bibliography{sample-base}

\end{document}